\documentclass[10pt,twocolumn,letterpaper]{article}

\usepackage{cvpr}
\usepackage{times}
\usepackage{epsfig}
\usepackage{graphicx}
\usepackage{amsmath}
\usepackage{amssymb}
\graphicspath{{media/}}

\cvprfinalcopy 

\def\cvprPaperID{****} 

\begin{document}

\title{Background subtraction based on Local Shape }
\author{Jean-Philippe Jodoin\\
\and
Guillaume-Alexandre Bilodeau\\
\and
Nicolas Saunier\\
\'Ecole Polytechnique de Montréal\\
P.O. Box 6079, Station Centre-ville, Montr\'eal, (Qu\'ebec), Canada, H3C 3A7\\
{\tt\small \{jean-philippe.jodoin, guillaume-alexandre.bilodeau, nicolas.saunier\}@polymtl.ca}
}

\maketitle

\begin{abstract}
We present a novel approach to background subtraction that is based on the local shape of small image regions. In our approach, an image region centered on a pixel is modeled using the local self-similarity descriptor. We aim at obtaining a reliable change detection based on local shape change in an image when foreground objects are moving. The method first builds a background model and compares the local self-similarities between the background model and the subsequent frames to distinguish background and foreground objects. Post-processing is then used to refine the boundaries of moving objects. Results show that this approach is promising as the foregrounds obtained are complete, although they often include shadows. 
\end{abstract}

\section{Introduction}
Background subtraction methods are an important step in numerous computer vision systems. These methods are used to identify moving objects in a video stream, which is often the first step in complex systems such as activity recognition, object tracking, and motion capture. Extracting the moving objects can improve the reliability of the system by reducing the search space, reducing processing needs, and allowing the use of simpler technics for the rest of the data extraction. Needless to say, the quality of many computer vision systems directly depend on the quality of the background subtraction method used. 

Most background subtraction methods work at the pixel level like the classic single Gaussian method~\cite{mckenna2000tracking} and the Gaussian mixture model~\cite{stauffer1999adaptive}. The shortcomings of these methods are that they may be affected by noise and perturbations in the image, as no notion of neighborhood consistency is used. This problem is difficult to solve at the pixel level and are why we developed a new local shape-based approach to background subtraction based on the Local Self-Similarity (LSS) descriptors~\cite{shechtman2007matching}.

Our approach is not unlike other region-based method like~\cite{varcheie2010bilodeau}, but in our case, we use the LSS descriptor to find the foreground regions instead of using histogram and the color intensity of the pixels inside rectangular regions. We also use a simple post-processing step to refine the objects' boundary accuracy, as the descriptors cover regions that are larger than a pixel. Our post-processing does not include the removal of shadows and we did not yet consider dynamic backgrounds.

\section{Methodology}
\subsection{Background model}
The first step in our method involves the creation of a background model. This model is a representation of the background with no foreground objects in it. The resulting model will be a grid of local self-similarity descriptors and a background image. To build this model, we use a set of training frames, in which we calculate the self-similarity descriptors using default parameters~\cite{2008horster} for each pixel. The descriptor, centered on the pixel is a log-polar representation of a correlation surface resulting from comparing with a sum of square differences a 5x5 pixel patch inside a 41x41 pixel patch. The correlation surface is expressed as an 80 components self-similarity vector (20 angles, 4 radial intervals)~\cite{stauffer1999adaptive}. For the subsequent frames in the training set after the processing of the first frame, we calculate the self-similarity descriptors for all pixels and we calculate the Euclidean distance to the existing region descriptor for all corresponding pixel positions. If the distance is below a threshold (a threshold of 1 was used), we increment a counter for that descriptor, otherwise we create a new descriptor for the region at this pixel and put the counter value to 1. This process is repeated as long as there are training frames available. If the camera is not static, or if the background is dynamic, there will be a lot of descriptors for a single image pixel. When all training frames are processed, the descriptor for a given pixel position with the highest count value (frequency) is selected as the background descriptor of the region. We assume that every pixel will represent more frequently the background than anything else in the training dataset. If a static foreground object is part of more than half of the training frames, it will be part of the background model. The final background model is composed of the LSS descriptors of each pixel and the pixel colors for the corresponding pixel. The pixel colors are kept for the post-processing part. The background model is static at the moment, but future research will aim to take into account dynamic lighting condition and intermittent motion from moving objects.

\subsection{Foreground object detection}
To detect  changes in a frame, we use a process similar to the one used for the creation of the background model. For a new frame, we calculate the Euclidean distance between each pixel descriptor and the corresponding background pixel descriptor, and if it is higher than a threshold (a threshold of 30 was used), the pixel is assumed to be part of the foreground. This gives a good estimate of the foreground object's position, but it tends to overestimate the size of the objects because of the way the LSS descriptor works. This is due to the fact that the local self-similarity descriptor correlates a 5x5 image patch with a larger surrounding region (41x41 pixel patch). Using too small region patch tends to make the descriptor less robust, so we kept the recommended parameters of the algorithm. For this same reason, we added padding to the frames to be able to do the correlation with the larger pixel patch and avoid losing information on the border. The padding is used in our method so that we can have a neutral effect on the correlation. We can calculate the size of the padding border to add to an image using 
\begin{equation}
Padding = b-b \% 3
\end{equation}
with
\begin{equation}
b = r + p
\end{equation}
where r is the radius around patch and p is the patch size.

\subsection{Post-processing}
\begin{figure}
\begin{center}
 \includegraphics[width=3in]{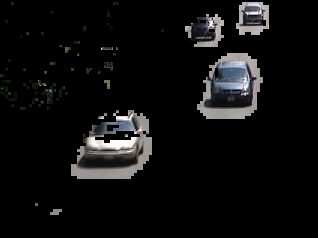}
\caption{LSS background subtraction without post-processing}
\label{fig:nopostprocess}
\end{center}
\end{figure}

\begin{figure}
\begin{center}
 \includegraphics[width=3in]{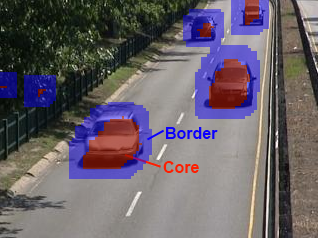}
\caption{Object core and border}
\label{fig:coreborder}
\end{center}
\end{figure}

The larger surrounding regions around the foreground objects reduce the precision of the method. There is also a small amount of noise due to dynamic background that changes the local self-similarity at a pixel. However, most of the foreground objects are complete. A result without any post-processing is shown in figure~\ref{fig:nopostprocess}.
To get more precise boundaries of objects, a series of morphological operations are applied. First, a closing is used to remove the holes in the foreground objects. After that, an erosion operation is performed on the foreground objects to remove as much noise as possible. Finally, a dilation technique is applied to the eroded objects, and subtracting the dilated objects from the eroded objects gives us an approximation of the boundaries of the objects. The dilation parameter should be adjusted in order to have a border with a size similar to the radius use in the local self-similarity calculation. The foreground objects are now separated in a core part and a border part (as shown in figure~\ref{fig:coreborder}). The core part will be used directly in the foreground mask and the border part will need further refinements. 

To refine the border of the foreground object, we use a simple Euclidean distance between the color intensity of the border pixels and the color intensity of the corresponding pixel in the background model. If the distance between the two pixels is over a threshold (the threshold used was 30), we consider the pixel to be part of the foreground, and else it is part of the background. 

\begin{figure}
\begin{center}
 \includegraphics[width=3in]{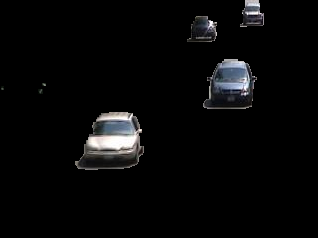}
\caption{LSS background subtraction after post-processing}
\label{fig:postprocess}
\end{center}
\end{figure}

The resulting mask is then eroded to remove the noise in the border and a closing is applied to have a cleaner and more precise foreground mask. The result of this step is shown in figure~\ref{fig:postprocess}.

\section{Results}
To compare our method to state-of-the-art methods, we have used the change detection datasets available from~\cite{goyette2012jodoin} and we have applied our method on three categories, one of them is the baseline dataset which is a scene with an almost static background and a static camera. We have also applied our method to the shadow dataset, and the thermal dataset which respectively contains a picture sequence with prominent shadows and thermal imagery. The cameraJitter dataset, the dynamicBackground dataset and the intermittent ObjectMotion dataset will not be covered by this method because those situations are not handled at the moment by the algorithm and they will be part of future work. For our data, we had four measures for each dataset, the number of true positive in the dataset (TP), the number of false positive (pixel detected as foreground that should have been detected as background) (FP), the number of false negative (pixel detected as background that should have been detected as foreground) (FN) and the number of true negative (TN).  With these, we have calculated the following metrics as defined in~\cite{goyette2012jodoin}:

\begin{equation}
Recall = \frac{TP}{TP+FN}
\end{equation}

\begin{equation}
Specifity = \frac{TN}{TP+FN}
\end{equation}

\begin{equation}
False Positive Rate = \frac{FP}{FP+TN}
\end{equation}

\begin{equation}
False NegativeRate = \frac{FN}{TN+FP}
\end{equation}

\begin{equation}
\% of Bad Classification = 100*\frac{FN+FP}{TP+FN+FP+TN}
\end{equation}

\begin{equation}
Precision = \frac{TP}{TP+FP}
\end{equation}

\begin{equation}
F-Measure = \frac{2*Precision*Recall}{Precision+Recall}\\
\end{equation}

To calculate the rank of the methods in the tables, we  calculated the rank for each method in each metric and the methods were sorted by the average rank of the metrics. The results of the other method are from~\cite{goyette2012jodoin}. The values in the result table are the average value across all datasets from a category. As shown in table~\ref{table:tab1}, the recall metric shows that our methods does not miss a lot of pixels from the moving objects. It has a higher rate of false positive than other methods, but it still achieves a reasonable percentage of bad classification (PBC).  

\begin{table}[h!]
\begin{center}
 \includegraphics[width=3in]{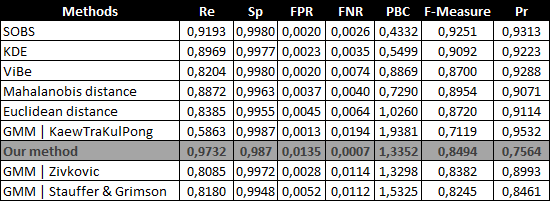}
\caption{Metrics for our method applied to the baseline dataset}
\label{table:tab1}
\end{center}
\end{table}

\begin{figure}[h!]
\begin{center}
 \includegraphics[width=3in]{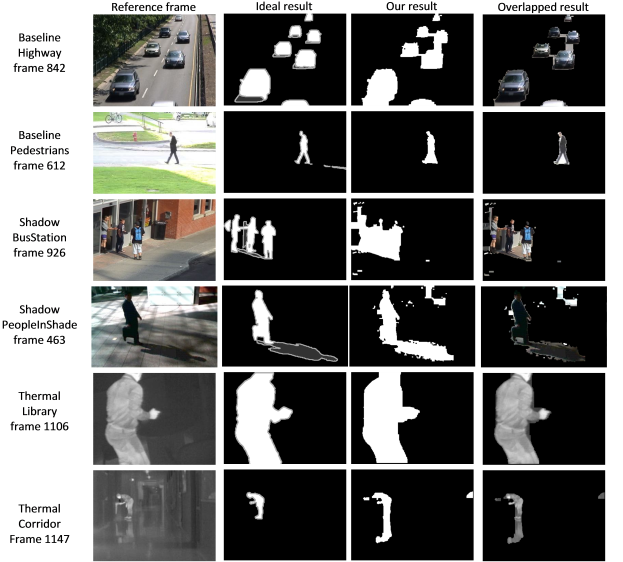}
\caption{Results for the baseline, shadow and thermal dataset}
\label{fig:results}
\end{center}
\end{figure}

The LSS descriptor is a good way to detect changes in images because moving objects result in a change in shape. The difficulty is refining the results to the pixel level. In this paper, the refinement is done with a simple Euclidean distance that is not adaptive, and simple morphological operations. Still, the method ranks reasonably well. Because shadows cause changes in the local correlation surface, they are systematically detected. Furthermore, small holes in objet (like the space between the legs, see the second row of figure~\ref{fig:results}) are included in the foreground because they are smaller than the correlation surface size. However, this is beneficial within objects because perturbations at the pixel level or smaller than the correlation surface do not affect the detection.   

\begin{table}
\begin{center}
 \includegraphics[width=3in]{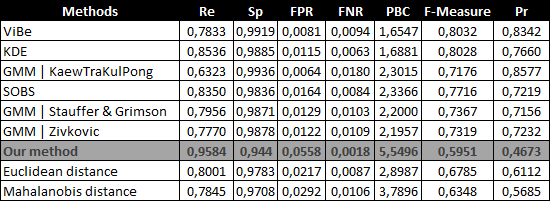}
\caption{Metrics for our method applied to the shadow dataset}
\label{table:tab2}
\end{center}
\end{table}

\begin{figure}
\begin{center}
 \includegraphics[width=3in]{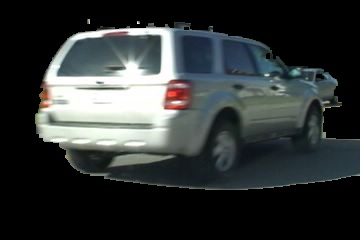}
\caption{Frame 362 of the shadow/bungalows dataset}
\label{fig:shadow}
\end{center}
\end{figure}

In table~\ref{table:tab2}, the false positive rate increases significantly compared to the baseline dataset. This is due to the detection of shadows as a new shape by the LSS descriptor. This effect is quite visible in figure~\ref{fig:shadow}. Shadows change the local correlation surface because details are less visible as the intensities gets darker. 

\begin{table}
\begin{center}
 \includegraphics[width=3in]{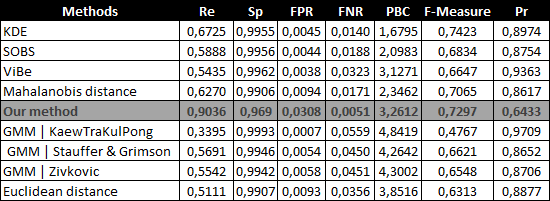}
\caption{Metrics for our method applied to the thermal dataset}
\label{table:tab3}
\end{center}
\end{table}

For the thermal dataset, our method had no problem to find all the moving parts and shows a high rate of recall. This is due to the fact that the moving objects (humans) boundaries are well defined in the thermal images. The thermal reflections of the humans are also very well defined as we can see in figure~\ref{fig:results}. The algorithm detects them as part of the body with an almost perfect symmetry. This explains the high level of false positive. A possible way to eliminate those reflections would be to combine the thermal camera in stereo with a visible camera, as it was already done by~\cite{torabi2011bilodeau} using LSS.

\section{Conclusion and future work}
In this paper, we have used the LSS descriptor as a way to distinguish foreground objects from the background. We have successfully built a static model of the background and used a metric to determine if the pixel patches were part of the foreground or the background. After that, we used the color information and morphological operation to refine the model border. The use of LSS patches instead of individual pixel intensity provides some robustness to camera noise and small intensity change which provides more complete foreground objects. As a future direction for this work, we will be working on making the algorithm more resistant to small camera viewpoint change, long term change in picture (such as a parked car moving on the background) and shadow removal.

{\small
\bibliographystyle{ieee}
\bibliography{biblio}
}

\end{document}